\documentstyle[colap,epsfig]{article}
\newcommand{\Sp}[1]{\mbox{}\vspace{#1mm}\\}

\title{Research on Architectures for Integrated Speech/Language
  Systems in Verbmobil}
\author{G\"unther G\"orz, Marcus Kesseler, J\"org Spilker, Hans Weber \\
  University of Erlangen-N\"urnberg\\
  IMMD (Computer Science) VIII --- Artificial Intelligence \\
  Am Weichselgarten 9 \\
  D-91058 ERLANGEN \\
  Email: {\tt \{goerz,kesseler,spilker,weber\}@informatik.uni-erlangen.de}
  }

\begin{document}
\maketitle

\begin{abstract}
The German joint research project Verbmobil (VM) aims at the
development of a speech to speech translation system.  This paper
reports on research done in our group which belongs to Verbmobil's
subproject on system architectures (TP15).  Our
specific research areas are the construction of parsers for
spontaneous speech, investigations in the parallelization of parsing
and to contribute to the development of a flexible communication
architecture with distributed control.  
\end{abstract}
\section{Introduction}
The German joint research project Verbmobil (VM)\footnote{This work
was funded by the German Federal Ministry for Research and Technology
(BMFT) in the framework of the Verbmobil Project under Grant BMFT 01
IV 101 H / 9. The responsibility for the contents of this study lies
with the authors.} aims at the development of a speech to speech
translation system.  This paper reports on research done in our group
which belongs to Verbmobil's subproject on system architectures
(TP15).  The task of this subproject is to provide basic research
results on incremental and interactive system architectures for the
VM research prototype and to demonstrate their feasibility in the
prototypical INTARC system.  Our specific research areas are the
construction of parsers for spontaneous speech, investigations in the
parallelization of parsing and to contribute to the development of a
flexible communication architecture with distributed control.  The
paper is organized as follows: Section 2 reports on the design and
implementation of an incremental interactive speech parser which
integrates statistics with a chart-parser employing a unification
grammar (UG) formalism.  Furthermore, results of experiments on the
interaction between the parser and a speech recognizer using
expectations, are reported.  In section 3 we present experiences with
a parallel version of the parser.  Section 4 deals with distributed
control in modular Natural Language/Speech (NLSP) systems.
\section{Design and Implementation of Incremental Interactive
Speech Parsers}
In a Left Right Incremental architecture (LRI), higher level modules
can work in parallel with lower level modules.  The obvious benefits
of such an arrangement are twofold: The system does not have to wait
for a speaker to stop talking and top-down constraints from higher
level to lower level modules can be used easily.  To achieve LRI
behavior the singular modules must fulfill the following
requirements:

Processing proceeds incrementally along the time axis
(``left to right''). 

Pieces of output have to be transferred to the next module
as soon as possible.

So far in INTARC-1.3 we have achieved an LRI style coupling of four
different modules: Word recognition module, syntactic parser,
semantic module and prosodic boundary module.
Our word recognition module is a modified Viterbi decoder, where two
changes in the algorithm design were made: We use only the forward
search pass, and whenever a final HMM state is reached for an active
word model, a corresponding word hypothesis is sent to the parser.
Hence backward search becomes a part of the parsing algorithm.
The LRI parsing algorithm is a modified active chart parser with an
agenda driven control mechanism.  The chart vertices correspond to
the frames of the signal representation. Edges correspond to word or
phrase hypotheses, being partial in the case of active edges.  A
parsing cycle corresponds to a new time point related to the
utterance. In every cycle a new vertex is created and new word
hypotheses ending at that time point are read and inserted into the
chart.  In one cycle, a backward search is performed to the beginning
of the utterance or to some designated time point in the past
constituting a starting point for grammatical analysis.  Search is
guided by a weighted linear combination of acoustic score, bigram
score, prosody score, grammar derivation score and grammatical
parsability. The 
search prodecure is a beam search implemented as an agenda access
mechanism.  The grammar is a probabilistic typed UG with separate
rules for pauses and other spontanous speech phemomena.
\subsection{Basic Objects}
In the following we use record notation to refer to subcomponents of
an object.  A chart vertex $V_t$ corresponds to frame number $t$.
Vertices have four lists with pointers to edges ending in and
starting in that vertex: {\em inactive-out, inactive-in, active-out}
and {\em active-out\/}.  A word hypothesis $W$ is a quadruple {\em
(from, to, key, score)\/} with {\em from\/} and {\em to\/} being the
start and end frames of $W$.  {\em W.Key\/} is the name of the
lexical entry of $W$ and {\em W.score\/} is the acoustic score of $W$
for the frames spanned, given by a corresponding HMM acoustic word
model. An edge $E$ consists of {\em from\/}, the start vertex and {\em
to\/}, a list of end vertices. Note that after a Viterbi forward pass
identical word hypotheses do always come in sequence, differing only
in ending time. {\em E.actual\/} is the last vertex added to {\em
E.to\/} in an operation.  Those ``families'' of hypotheses are
represented as one edge with a set of end vertices.  {\em E.words\/}
keeps the covered string of word hypotheses while SCORE is a record
keeping score components.  Besides that an edge consists of a
grammar rule {\em E.rule\/} and {\em E.next\/}, a pointer to some
element of the right hand side of {\em E.rule\/} or {\em NIL\/}.  As
in standard active chart parsing an edge is passive, if {\em E.next =
nil\/}, otherwise it is active.  {\em E.cat\/} points to the left
hand side of the grammar rule.
SCORE is a record with entries for inside and outside probabilities
given to an edge by acoustic, bigram, prosody and grammar model:
\begin{description}
\item[Inside-X] Model scores for the spanned portion of an edge.
\item[Outside-X] Optimistic estimates for the portion from vertex 0 to
the beginning of an edge.
\end{description}
For every vertex we keep a best first store of scored edge pairs. We
call that store $\mbox{Agenda}_i$ in cycle $i$. 
\subsection{Basic Operations}
There are five basic operations to define the operations of the
parsing algorithm.  The two operations {\em Combine\/} and {\em Seek
Down\/} are similar to the well known Earley algorithm operations
{\em Completer\/} and {\em Predictor\/}.  Furthermore, there are two
operations to insert new word hypotheses, {\em Insert\/} and
{\em Inherit\/}.  All these operations can create new edges, so
operations to calculate new scores from old ones are attached to
them.  In order to implement our beam search method appropriately but
simply, we define an operation {\em Agenda-Push\/}, which selects
pairs of active and passive edges to be pruned or to be processed in
the future. The basic operations are given in CFG notation for
simplicity.
\subsubsection{Combine}
For a pair of active and passive edges $(A,I)$, if {\em A.next =
I.cat\/} and {\em I.from\/} $\in$ {\em A.to\/}, insert edge $E$ with
{\em E.rule = A.rule, E.cat = A.cat, E.next = shift(A.next), E.from =
A.from, E.to = A.to\/}. \\
For $X$ = Bigram, Grammar and Prosody:\\ 
{\em E.Outside-X = A.Outside-X + I.Inside-X + Trans(X,A,I)\\
E.Inside-X = A.Inside-X + I.Inside-X + Trans(X,A,I)\/} \\
For $X$ = Acoustic:\\
{\em E.Outside-X = A.Outside-X[I.from] $\oplus$  I.Inside-X  Trans(X,A,I)\\
E.Inside-X = A.Inside-X[I.from] $\oplus$  I.Inside-X  Trans(X,A,I)\/}\\
The operator  $\oplus$ performs an addition of a number to every
element of a set.
{\em Trans(X,A,I)\/} is the specific transition penalty a model will
give to two edges. In the case of acoustic scores, the penalty is
always zero and can be neglected.  In the bigram case it will be the
transition from the last word covered by $A$ to the first word covered
by $B$.
\subsubsection{Seek Down}
Whenever an active edge $A$ is inserted, insert an edge $E$ for
every rule $R$ such that {\em A.next = E.cat, E.rule = R, E.from = A.actual,
E.to = \{A.actual\} \/}. 
For $X$ = Acoustic, Prosody and Bigram:\\
{\em E.Inside-X = 0 \\
E.Outside-X = A.Outside-X \/}\\
For $X$ = Grammar: \\
{\em E.Inside-X = grammar score of R \\
E.Outside-X = A.Outside-X + Trans(X,A,E) + E.Inside-X \/}.
This recursive operation of introducing new active edges is
precompiled in our parser and extremely efficient.
\subsubsection{Insert}
For a new word hypothesis {\em W = (a,i,key,score)\/} such that no
{\em W' = (a,i-1,key,score')\/} exists, insert an edge $E$ with {\em
E.rule = lex(key), E.cat = lex(key), E.from = $V_a$, E.to = \{$V_i$\}
\/} and for $X$ = Acoustic:\\
{\em E.Inside-X = E.Outside-X = \{(i,score)\} \/},\\
for $X$ = Prosody and Bigram:\\
{\em E.Inside-X = E.Outside-X = 0,\\ for X = Grammar E.Inside-X =
E.Outside-X = grammar score of lex(key)\/}.
\subsubsection{Inherit}
For a new word hypothesis {\em W = (a,i,key,score)\/} such that a
{\em W' = (a,i-1,key,score')\/} exists: \\
For all $E$ in {\em $V_{i-1}$.inactive-in\/} or $V_{i-1}$.{\em
active-in\/}: If {\em last(E.words) = key\/} then add {\em
\{$V_i$\}\/} to {\em E.to\/}, add {\em (i,E.Inside-Acoustic[i-1] -
score' + score)\/} to {\em E.Inside-Acoustic\/} and add {\em
(i,E.Outside-Acoustic[i-1] - score' + score)\/} to {\em
E.Outside-Acoustic\/}.\\ If $E$ is active, perform a {\em
Seek-Down\/} on $E$ in $V_i$.
\subsubsection{Agenda Push}
Whenever an edge $E$ is inserted into the chart, if $E$ is active
then for all passive $A$, such that {\em A.from $\in$ E.to\/} and
{\em combined-score(E,A) $>$ Beam-Value\/}, insert {\em
(E,A,combined-score(E,A))\/} into the actual agenda.  If $E$ is
passive then for all active $A$, such that {\em E.from $\in$ A.to\/}
and {\em combined-score(A,E) $>$ Beam-Value\/}, insert {\em
(A,E,combined-score(A,E))\/} into the actual agenda.
{\em Combined-Score\/} is a linear combination of the outside
components of an edge $C$ which would be created by $A$ and $E$ in a
{\em Combine\/} operation.
{\em Beam-Value\/} is calculated as a fixed offset from the maximum
{\em Combined-Score\/} on an agenda. Since we process best-first
inside the beam, the maximum is known when the first triple is
inserted into an agenda. 
{\em Agenda-Pop\/} will remove the best triple from an actual agenda
and return it.
\subsection{A simple LRI lattice parser}
The follwing control loop implements a simple LRI lattice parser. 
\begin{enumerate}
\item $T = 0$. Create $V_T$
\item Insert initial active edge $E$ into $V_T$, with {\em E.next = S\/}
\item Increment $T$. Create $V_T$ \label{xxx}
\item For every $W$ with {\em W.end = T: Insert(W)\/} or {\em Inherit(W)\/}
\item Until {\em Agenda[T]\/} is empty:
  \begin{enumerate}
  \item {\em Combine(Agenda-Pop)\/}
  \item When combination with initial edge is successful, send
    result to SEMANTICS
  \end{enumerate}
\item Communicate with PROSODY and go to \ref{xxx}
\end{enumerate}
\subsection{The Grammar Model}
The UG used in our experiments consists of 700 lexical entries and 60
rules.  We used a variant of inside-outside training to estimate a model of
UG derivations. It is a rule bigram model similar to PCFG with
special extensions for UG type operations. The probability of future
unifications is made dependent from the result type of earlier
unifications.  The model is described in more detail in
\cite{weber:94a,weber:94b}; it is very similar to \cite{brew:95}.
\subsection{LRI Coupling with Prosody}
In INTARC we
use three classes of boundaries, B0 (no boundary), B2 (phrase
boundary), B3 (sentence boundary) and B9 (real break).  The prosody
module, developed at the University of Bonn, classifies time intervals
according to these classes. A prosody hypothesis consists of a
beginning and ending time and model probabilities for the boundary
types which sum up to one.
A prosodic transition
penalty used in the {\em Combine\/} operation was taken to be the
score of the best combination of bottom-up boundary hypothesis {\em
Bx\/} and a trigram score {\em (lword, Bx, rword)\/}. Here {\em
lword\/} is the last word of the edge to the left and {\em rword\/}
is the first word spanned by the edge to the right.
Prosody hypotheses are consumed by the parser in every cycle and
represented as attributes of vertices which fall inside a prosodic
time interval.
In a couple of tests we already achieved a reduction of edges of
about 10\% without change in recognition rate using a very simple
trigram with only five word categories.
\subsection{Experimental Results}
In a system like INTARC-1.3, the analysis tree is of much higher
importance than the recovered string; for the goal of speech
translation an adequate semantic representation for a string with
word errors is more important than a good string with a wrong reading.
The grammar scores have only indirect influence on the string;
their main function is picking the right tree.  We cannot measure
something like a ``tree recognition rate'' or ``rule accuracy'',
because there is no treebank for our grammar.
The word accuracy results cannot be compared
to word accuracy as usually applied to an acoustic decoder in
isolation.  We counted only those words as recognized which could be
built into a valid parse from the beginning of the utterance.  Words
to the right which could not be integrated into a parse, were counted
as deletions --- although they might have been correct in standard
word accuracy terms.  This evaluation method is much harder than
standard word accuracy, but it appears to be a good approximation to
``rule accuracy''.
Using this strict method we achieved a word accuracy of 47\%, which is
quite promising. 

Results using top down prediction of possible word hypotheses by the
parser -- work inspired by \cite{kita:89} --  have already been
published in \cite{hauenstein:94a,hauenstein:94b}, \cite{weber:94a},
and \cite{weber:94b}. Recognition rates had been improved there for
read speech. In spontaneous speech we could not achieve the same
effects. 
\subsection{Current Work}
Our current work, which led to INTARC-2.0, uses a new approach for
the interaction of syntax and semantics and a revision of the interaction of
the parser with a new decoder.  For the last case we implemented a precompiler
for word-based prediction which to our current experience is clearly
superior to the previous word-class based prediction.  For the implementation
of the interaction of syntax and semantics we proceed as follows: A new
turn-based UG has been written, for which a context-sensitive stochastic
training is being performed.  The resulting grammar is then stripped down to a
pure type skeleton which is actually being used for syntactic parsing.  Using
full structure sharing in the syntactic chart, which contains only packed
edges, we achieve a complexity of $O(n^3)$.  In contrast to that, for semantic
analysis a second, unpacked chart is used, whose edges are provided by an
unpacker module which is the interface between the two analysis levels.  The
unpacker, which has exponential complexity, selects only the $n$ best scored
packed edges, where $n$ is a parameter.  Only if semantic analysis fails it
requests further edges from the unpacker.  In this way, the computational
effort on the whole is kept as low as possible.
\section{Parallel Parsing}
One of our main research interests has been the exploration of
performance gains in NLP through parallelization. To this end, we
developed a parallel version of the INTARC parser. Although the
results so far are yet not as encouraging as we expected, our efforts
make for interesting lessons in software engineering.
The parallel parser had to obey the following restrictions: Running on our
local shared memory multiprocessor (SparcServer1000) with 6 processors,
parallelization should be controlled by inserting Solaris-2.4 thread and
process control primitives directly into the code.  The only realistic choice
we had was to translate our parser with Chestnut Inc.'s Lisp-to-C-Translator
{\em automatically\/} into C.  Since the Lisp functions library is available
in C source, we could insert the necessary Solaris parallelisation and
synchronization primitives into key positions of the involved functions.
\subsection{Parallelization Strategy and Preliminary Results}
For effective parallelization it is crucial to keep communication
between processors to a minimum.  Early experiments with a fully
distributed chart showed that the effort required to keep the
partial charts consistent was much larger that the potential gains of
increased parallelism.  The chart must be kept as a single data
structure in a shared memory processor, where concurrent reads are
possible and only concurrent writes have to be serialized with
synchronisation primitives.
An analysis of profiling data shows that even the heavily optimized
UG formalism causes between 50\% -and 70\% of the
computational load in the serial case.  Therefore we provide an
arbitrary number of {\em unification workers\/} running in parallel
which are fed unification tasks from the top of an agenda sorted by
scores.
Due to the high optimization level of the sequential parser,
load-balancing is fairly poor. Namely, the very fast type check used
to circumvent most unifications, causes large disparities in the
granularity of agenda tasks. Furthermore, pathological examples have
been found in which a single unification takes much longer than all
other tasks combined.
\begin{figure}[hbtp]\begin{center}\Sp{-5}%
\mbox{\epsfxsize=80mm\epsfbox{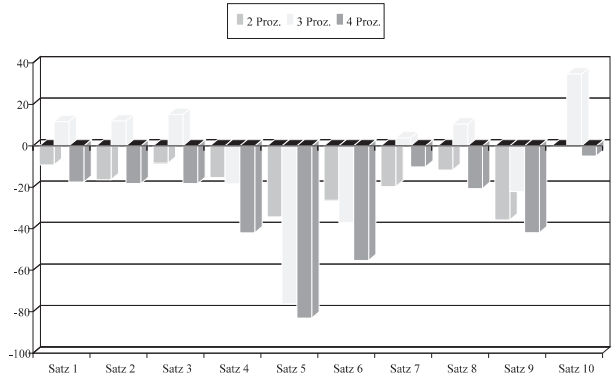}}
\end{center}
\caption{Percentual gains and losses over attained
over 10 different sentences (Spilker 1995)}
\label{fig:results}
\end{figure}
\section{Distributed Control in Verbmobil}
The question of control in VM is tightly knit with the architecture of the VM
system. As yet, the
concept of architecture in VM has been used mostly to describe the overall
modularization and the interfaces implied by the data flow between modules.
This socalled {\em domain architecture} is incomplete in the sense that it
does not specify any {\em interaction strategies\/}.
Within our research on interactive system architectures we developed
 a modular communication framework, ICE\footnote{based on PVM (parallel
virtual machine)}, in cooperation with the University of Hamburg.  Now, ICE is
the architectural framework of the VM research prototype.
\subsection{The INTARC Architecture}
\begin{figure}[hbtp]\begin{center}\Sp{-5}%
\mbox{\epsfxsize=70mm\epsfbox{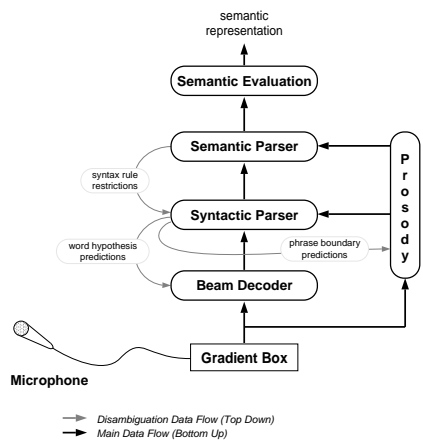}}
\end{center}
\caption{The interactive, incremental INTARC-1.3 architecture}
\label{fig:intarc}
\end{figure}

The INTARC architecture as first presented by \cite{Pyka:92d} is a distributed
software system that allows for the interconnection of NLSP modules under the
principles of incrementality and interactivity.  Figure~\ref{fig:intarc} shows
the modularization of INTARC-1.3: There is a main broad channel connecting all
modules in bottom-up direction, i.e., from signal to interpretation.
Furthermore, there are smaller channels connecting several modules, which are
used for the top-down interactive disambiguation data flow.  Incrementality is
required for all modules.
ICE assumes that each module has a local memory that is not directly
accessible to other modules.  Modules communicate explicitly with one another
via messages sent over bidirectional channels.  This kind of communication
architecture is hardly new and confronts us directly with a large number of
unresolved issues in distributed problem solving, cf.\ \cite{Durfee:89}.  In
the last 20 years there have been numerous architecture proposals for
distributed problem solving among computing entities that exchange information
explicitly via message passing.  None of these models include
explicit strategies or paradigms to tackle the problem of distributed control.
\subsection{Structural Constraints of Verbmobil}
\label{sec:dc} Modularity, being a fundamental assumption in VM
\cite{Wahlster:92}, does still leave us with two problems: First,
modules have to 
communicate with one another, and second, their local behaviors have
to be somehow coordinated into a coherent global, possibly optimal,
behavior.  Unfortunately, the task of system integration has to obey
some {\em structural constraints\/} which are mostly pragmatic
in nature:
\begin{itemize}
\item Some of the modules are very complex software systems in
  themselves.  Highly parameterizable and with control subtly
  spread over many interacting submodules, understanding and then
  integrating such systems into a common control strategy can
  be a very daunting task.
\item Control issues are often very tightly knit with the domain
  the module is aimed at, i.e., it is very difficult to understand
  the control strategies used without sound knowledge of the
  underlying domain.  The problem even gets
  worse if what is to be fine-tuned is the {\em interaction\/}
  between several complex modules.
\end{itemize}
These two arguments are similar in nature, but differ in the
architectural levels that they apply to. The former is implementation
related, the latter algorithm and theory related.
\subsection{Layers of Control}
Modules have to communicate with one another and their local
behaviors have to be coordinated into a coherent global, possibly
optimal, behavior.  In highly distributed systems we generally find
the following levels of control:

\paragraph{System Control:} 
The minimal set of operating system related actions that each
participating module must be able to perform which will
typically include means to start up, reset, monitor, trace
and terminate individual modules or the system as a whole.
\paragraph{Isolated Local Control:} 
The control strategies used within the module disregarding any
interactions beyond initial input of data and final output of
solutions. There is only one thread of control active at any time. 
\paragraph{Interactive Local Control:} 
Roughly, this can be seen as isolated local control extended with
interaction capabilities.  
{\em Incrementality\/} is given by the possibility of control flowing back
to 
a certain internal state after an output operation.
Higher {\em interactivity\/} is made possible by entering
a state more often from various points within the module and by
adding a new waiting loop 
to check for any top-down requests.  The requirement for anytime behavior 
is a special case of that \cite{goerz:94}.

In our experience the change to interactive control will tremendously
increase the complexity of the resulting code.  But we are still
making the simplifying assumptions that the algorithm 
can be used incrementally --- but there are
algorithms unsuitable for incremental processing (e.g.\ $A^*$).
Incrementality can lead to the demand for a complete redesign of a module.
Furthermore we assume that simply by exchanging data and doing simple
extensions in the control flow everything will balance out nicely on
the system scale which is enormously naive. Even for 
the sequential architecture implied by the case of isolated local control,
we have to solve a whole plethora of new problems that come along with
interactivity:
\begin{itemize} 
\item Mutual deadlock
\item Mutual live-lock
\item Race conditions (missing synchronization)
\item Over-synchronization
\end{itemize}
\paragraph{Dialogue Control:} 
In systems like VM there {\em is\/} a module that comes
close to possessing the ``integrated view'' of a centralized
blackboard control: the {\em dialogue module\/}.  So it seems the
right place to handle some of the global strategic control issues,
like:
\begin{itemize}
\item Domain error handling 
\item Observe timeout constraints 
\item Resolve external ambiguities/unknowns 
\end{itemize}
The fact that the dialogue module exercises a kind of {\em global
control\/} does not invalidate what has been said about the
unfeasability of central control, because the control exercised by it
is very coarse grained.  To handle finer grained control
issues in any module would take us back to memory and/or
communication system contention.

\end{document}